\documentclass[preprint,11pt]{elsarticle}
\usepackage{enumitem}
\usepackage{amssymb}
\usepackage{eurosym}
\usepackage{color}
\usepackage{bm}
\usepackage{booktabs}
\usepackage{enumitem}
\usepackage{multirow}
\usepackage{longtable}
\usepackage{arydshln}
\usepackage[top=2.2cm,bottom=2.5cm,left=2.5cm,right=2.5cm]{geometry}
\usepackage{hyperref}
\usepackage{mathtools}
\usepackage{algorithm,algpseudocode}
\usepackage{subcaption}
\usepackage{tikz}
\usepackage{pgfplots}
\usetikzlibrary{patterns}
\usetikzlibrary{arrows}
\usepackage{amsmath}
\usepackage{etoolbox}
\usepackage{setspace}
\usepackage{etoolbox}

\usepackage{amsthm,amsmath,eurosym}

\newcommand{\R}{\mathbb{R}}

\newcommand{\pt}{\text{ }\forall\text{ }}

\newcommand{\N}{\mathbb{N}}

\newcommand{\nint}[2]{[#1,#2]\cap\N}

\pgfplotsset{compat=1.18} 

\journal{European Journal of Operational Research}

\usepackage{lineno, blindtext}

\makeatletter
\def\ps@pprintTitle{%
  \let\@oddhead\@empty
  \let\@evenhead\@empty
  \def\@oddfoot{\reset@font\hfil\thepage\hfil}
  \let\@evenfoot\@oddfoot
}
\makeatother

\newtheorem{defi}{Definition}
\newtheorem{prop}{Proposition}

\newtheorem{exam}{Example}
\newtheorem{theo}{Theorem}
\newtheorem{cor}{Corollary}
\newtheorem{lem}{Lemma}
\begin{document}

\begin{frontmatter}
    \title{The Tournament Tree Method for preference elicitation in Multi-criteria decision-making}

    \author[JAEN1,cor]{Diego {\sc  Garc{\'i}a-Zamora}}\ead{dgzamora@ujaen.es}
   \author[JAEN2]{\'Alvaro {\sc Labella}}\ead{alabella@ujaen.es}
    \author[CEGIST]{José {\sc Rui~Figueira}}\ead{figueira@tecnico.ulisboa.pt }
    
    \address[JAEN1]{Department of Mathematics, Universidad de Ja{\'e}n, 23071 Ja{\'e}n, Spain}
    \address[JAEN2]{Department of Computer Science, Universidad de Ja{\'e}n, 23071 Ja{\'e}n, Spain}
    \address[CEGIST]{CEGIST, Instituto Superior T\'{e}cnico,  Universidade de Lisboa, Portugal}
    \cortext[cor]{Corresponding author at: Department of Mathematics, University of Ja{\'e}n, 23071 Ja{\'e}n, Spain}
    \begin{abstract}
    \noindent Pairwise comparison methods, such as Fuzzy Preference Relations and Saaty's Multiplicative Preference Relations, are widely used to model expert judgments in multi-criteria decision-making. However, their application is limited by the high cognitive load required to complete $m(m-1)/2$ comparisons, the risk of inconsistency, and the computational complexity of deriving consistent value scales. This paper proposes the Tournament Tree Method (TTM), a novel elicitation and evaluation framework that overcomes these limitations. The TTM requires only $m-1$ pairwise comparisons to obtain a complete, reciprocal, and consistent comparison matrix. The method consists of three phases: (i) elicitation of expert judgments using a reduced set of targeted comparisons, (ii) construction of the consistent pairwise comparison matrix, and (iii) derivation of a global value scale from the resulting matrix. The proposed approach ensures consistency by design, minimizes cognitive effort, and reduces the dimensionality of preference modeling from $m(m-1)/2$ to $m$ parameters. Furthermore, it is compatible with the classical Deck of Cards method, and thus it can handle interval and ratio scales. We have also developed a web-based tool that demonstrates its practical applicability in real decision-making scenarios.
    \end{abstract}
    \vspace{0.25cm}
\end{frontmatter}

\vfill\newpage

\section{Introduction}

\noindent The use of pairwise comparisons to model experts' opinions in decision-making problems has been extensively studied in the specialized literature \cite{Sahoo}. This method is preferred because humans generally perform better when eliciting preferences through pairwise comparisons among a set of objects (alternatives, criteria, scale levels,...) rather than rating each object individually \cite{Jensen1984,Miller1956}. In decision-making literature, two of the most widely-used types of pairwise comparison matrices are Fuzzy Preference Relations (FPRs) \cite{Orlovski1978}, and Multiplicative Preference Relations (MPRs) \cite{Saaty1990}. 

However, completing a pairwise comparison matrix imposes a significant time burden and cognitive challenge on decision-makers due to its inherent complexity and the meticulous nature of the task \cite{REZAEI201549}. Evaluating each object individually against each other requires extensive deliberation and attention to detail \cite{Jensen1984}. This process requires careful consideration of nuanced differences in preferences, which can be mentally taxing and prone to subjective interpretation. Decision-makers must manage a large volume of pairwise comparisons, increasing the likelihood of cognitive fatigue and potential errors in judgments \cite{Miller1956}. Moreover, the task's iterative nature necessitates repeated mental effort, which can further exacerbate decision fatigue and diminish overall decision quality. 

For this reason, inconsistencies can arise during the elicitation \cite{Saaty2008}, leading to contradictory evaluations that can significantly alter the decision-making process if left unaddressed \cite{REZAEI201549,Saaty2008}. To manage inconsistent preference relations, two primary strategies emerge. The first involves prompting experts to revise their preferences until achieving sufficient consistency \cite{REZAEI201549}. Alternatively, an automated mechanism can be employed to adjust the matrix until it achieves consistency \cite{XUinfsc}. However, both methods have drawbacks. The former can be highly time-consuming or even infeasible if the expert is no longer available to revise their preferences. The latter method automatically modifies the preferences, potentially altering the original opinions provided by the expert.
    
Furthermore, obtaining an overall performance score (or value scale) for the objects from a pairwise comparison matrix is not a trivial task. Each element in such a matrix represents the intensity of preference of one object over another on a certain scale, which requires sophisticated technical methods to synthesize these pairwise comparisons into a comprehensive value scale \cite{corrente2021}. This process must account for the inherent fuzziness and potential inconsistencies in the preferences, necessitating advanced computational techniques and algorithms. In addition, ensuring that the final value scale accurately reflects the original fuzzy judgments involves resolving any inconsistencies and validating the results, further complicating the process \cite{LIANG2020102175}. 

Another significant shortcoming of pairwise comparison matrices is their high computational complexity due to dimensional differences \cite{Rodriguez2021}. A pairwise comparison matrix to rate $m$ objects must be represented as an \( m \times m \) matrix, requiring \( m^2 \) entries to capture all pairwise comparisons. This is in stark contrast to a utility vector, which only needs \( m \) entries. The quadratic increase in the number of entries leads to a substantial increase in the computational effort needed for processing, aggregating, and ensuring consistency within the matrix. As the number of objects grows, the computational burden exponentially increases, making pairwise preference relations a resource-intensive and challenging tool for decision-making, not only during the elicitation process. 

To overcome these limitations, this paper proposes a new elicitation methodology based on pairwise comparisons. Such a methodology, the so-called Tournament Tree Method (TTM), is inspired by the development of a sports tournament and consists of asking a few specific questions about the objects that mathematically provide the necessary information to retrieve a complete, consistent pairwise comparison matrix \cite{AGOSTON2022102576}. The TTM evaluates $m$ objects through three steps. First, a questioning phase is carried out, in which only $m-1$ questions about the objects are asked to the decision-maker. In each question, the decision-maker is asked to compare a specific pair of objects by pointing out which one is the most preferred and which is the corresponding preference difference of attractiveness between them, which will be modelled by the introduction of blank cards as in the Deck of Cards Method \cite{corrente2021}. Second, a construction phase is developed to obtain a complete $m\times m$ pairwise comparison matrix which is reciprocal and consistent. Finally, an evaluation phase is included to derive $m$ global scores for the objects from such a pairwise comparison matrix. As a key feature of the method, these $m$ scores not only summarize the information within the pairwise comparison matrix, but also allow the reconstruction of the complete  $m\times m$ pairwise comparison matrix. 

As a result, we develop a methodology with the following advantages: (i) It has a low cognitive burden for the decision-makers, (it is much faster than asking for a complete pairwise comparison matrix), (ii) it allows constructing a consistent pairwise comparison matrix as well as deriving an easy-to-compute value scale/priority vector that contains all the information within the matrix, (iii) it enables representing the decision-maker's preference in a computationally efficient way since such a preference is modeled through $m$ values, instead of $m(m-1)/2$,  and, (iv) it is compatible with interval and ratio scales.    

The remainder of this paper is as follows. In Section \ref{sec:prel} presents the previous research that supports our proposal. Section \ref{sec:ttm} develops our proposed methodology in detail, providing the theoretical results that ensure its good performance. Section \ref{sec:app}, shows an application example that is carried out using the web-based tool that we have developed to facilitate the implementation of our method in real-world scenarios. Finally, Section \ref{sec:conclusion} concludes the paper.

%
\section{Literature review}\label{sec:prel}
\noindent In expert-driven decision-making, eliciting experts' opinions about objects is challenging because the verbalized values provided by them often do not truly reflect their actual thoughts \cite{Jensen1984,Miller1956}. To enhance decision quality, the concept of a pairwise comparison matrix was introduced to better capture experts' preferences \cite{Orlovski1978}. Unlike rating objects separately, pairwise comparisons require experts to express their preference for one object over another, one pair at a time. After completing the pairwise comparisons, an auxiliary method is applied to determine the priorities of each object. Prominent methods for this purpose include the eigenvalue method \cite{Saaty2008}, the Best-Worst Method \cite{REZAEI201549,REZAEI2016126}, dominance computation, or the Deck of Cards method \cite{corrente2021,LIANG2025103224}.

Pairwise comparisons are usually summarized in a preference table (or matrix) $M$, for which each item $M_{ij}$ contains information about the difference of attractiveness between two objects, which is used to compute the intensity of the $i$-th object over the $j$-th one. That preference intensity can be measured in either an interval scale \cite{corrente2021} or in a ratio scale \cite{Saaty1990}. Depending on the scale, the preference matrix presents a different kind of reciprocity condition, i.e., the relationship between $M_{ij}$ and $M_{ji}$. Additionally, according to the scale, a consistency condition is considered to recognize when the preference matrix presents contradictions within itself \cite{IVPR}.

In this paper, we consider the preference tables studied in \cite{corrente2021}, which have been shown to generalize other common preference matrices in the literature \cite{IVPR}. Formally, we work with matrices $M\in\mathcal{M}_n(\R)$ for which the element $M_{ij}$ represents the number of units that represent the difference of attractiveness between objects $i$ and $j$. When the difference of attractiveness of $i$ over $j$, $M_{ij}$ is a positive number, whereas if $j$ is preferred to $i$, $M_{ij}$ is a negative number. $M_{ij}=0$ means that there is no difference between the objects $i$ and $j$. Initially, we do not assume that the matrix is further normalized, though we will discuss how to do it later. 

As in \cite{corrente2021}, we assume that the number of units between objects is elicited using the Deck of Cards method. The decision-maker will be asked to place blank cards between the objects $i$ and $j$, according to the difference in attractiveness that he/she feels between such levels. The sign of $M_{ij}$ is determined by asking which object is more preferred, whereas its value will be one plus the number of cards he/she has placed. Remember that $0$ cards do not mean indifference, but that the difference {of attractiveness between the two objects} is minimal (equal to the unit) \cite{corrente2021}. Finally, the reciprocity condition for these preference matrices will be given by $M_{ij}+M_{ji}=0$, whereas the consistency condition is $M_{ik}+M_{kj}=M_{ij} \pt i,j,k=1,...,n$ \cite{IVPR}.

\section{The Tournament Tree Method}\label{sec:ttm}
\noindent This section is devoted to developing our preference elicitation methodology. Such a method presents three phases: Tournament, Processing, and Evaluation, which cover the interaction with the decision-maker, the construction of a pairwise comparison matrix, and the final computation of the ratings for the objects.

\subsection{Stage I. Tournament}
\noindent In the first step of this method, the decision-maker is asked to make some strategic pairwise comparisons between the corresponding objects and place blank cards between them to model the more or less difference in attractiveness between the two objects. Motivated by the scheme of most sports tournaments, each one of these comparisons will be a match between the corresponding pair of objects. Consequently, from each match we get one winner, one loser, and the score of the winner against the loser. After all the matches in one round are finished, only the winners go to the next round, in which they will face new opponents until only one object remains.  From the interaction point of view, the decision-maker will be asked to use blank cards to represent the difference in attractiveness between the most preferred object (winner) and the least preferred object. This process is repeated for specific pairs of objects across different rounds until the winner of the tournament is determined. Let us describe this process step by step.

\begin{itemize}
    \item[~]Step 1. Consider $m_r\in\N\setminus\{1\}$ objects that have to be evaluated in the round $r$. 
    \item[~] Step 2. The objects are grouped as pairs. If there is an odd number of objects, one of the objects is (virtually) doubled and paired with itself.
    \item[~] Step 3. For each pair, the decision-maker selects the best and worst objects and rates how much he/she prefer the best object over the worst one. This difference of attractiveness will be obtained using the Deck of Cards method: the decision-maker places cards according to his/her feelings, and then one unit is added to the number of cards placed. Keep in mind that $0$ cards do not represent indifference, but that the difference is equal to the unit (minimal)\cite{corrente2021}. At this moment, we store the winner, the loser, and the result of this match, i.e., the number of units representing the difference in attractiveness between the objects. In case one object is paired with itself, it is not compared, and it is considered victorious.
    \item[~] Step 4. The victorious objects in Step 3 move forward to the next round.
    \item[~] Step 5. Redefine the set of objects as the remaining ones, which contains $m_{r+1}$ elements.
    \item[~] Step 6. If $m_{r+1}=1$, the process finishes. Otherwise, go back to Step 2.
\end{itemize}

This process of assessing objects via a tournament will be called TTM. A decision-making problem consisting of evaluating objects through the TTM will be called a TT problem. Below, we show some theoretical results that bring insights into this methodology.

\begin{prop}
    Consider a round $r\in\N$ of a TT problem in which $m_r\in\N$ objects must be compared. Then, $m_{r+1}=m_r-F(m_r/2)$,  where $F$ denotes the floor function.   
\end{prop}
\begin{proof}
    Note that the number of pairwise comparisons that will be performed is $F(m_r/2)$, which means that $F(m_r/2)$ will be dropped in the next Step. Consequently, the number of remaining objects is $m_{r+1}=m_r-F(m_r/2)$.
\end{proof}

\begin{theo}
    Consider a TT problem in which $m\in\N$ objects must be compared. Then, the total number of comparisons performed in all the rounds is $m-1$ and the number of rounds is $R=C(\log_2(m))$, where $C$ denotes the ceil function.
\end{theo}
\begin{proof}
    First, note that during the TTM, we drop an object if and only if we have compared it and it has lost. At the end of the process, only one object remains, which means that we have dropped $m-1$ objects and, therefore, the decision-maker has performed $m-1$ comparisons. On the other hand, notice that for $m\in\N$ we can find $n\in \N$ satisfying
    \begin{equation*}
        2^{n}\leq m < 2^{n+1}.
    \end{equation*}
    So, assume two TT problems with $2^{n}$ and $2^{n+1}$ objects to be assessed, respectively. For each round that we carry out, we divide the number of objects by $2$ in both TT problems. Therefore, after $n$ rounds, we have:
    \begin{equation*}
        \frac{2^{n}}{2^{n}}\leq \frac{m}{2^{n}} < \frac{2^{n+1}}{2^{n}} \iff 1\leq \frac{m}{2^{n}}<2.
    \end{equation*}
    This means that to assess $m$ objects, we need to carry out either $n$ rounds, if $m=2^n$, or $n+1$ rounds if $m>2^n$. Since $n\leq \log_2(m)<n+1$, 
    \begin{equation*}
        R=C(\log_2(m)).
    \end{equation*}
\end{proof}

    \begin{prop}
        Let us consider a TT problem in which $m\in\N$ objects must be compared and consider $ R=C(\log_2(m))$. Then, in the round $r\in\nint{1}{R}$ the decision-maker performs at most $2^{R-r}$ comparisons. In addition, the total number of virtual comparisons is $2^R-m$.
    \end{prop}
    \begin{proof}
    The first statement is clear. In the last round ($R$), the decision-maker performs one comparison $2^{R-R}$, in the previous the double, and so on. Since we carry out $R$ rounds, the maximum number of comparisons will be $2^{R-1}$ because originally we have, at most, $2^{C(\log_2(m))}$ objects to compare.
    Now, note that the number of virtual comparisons is given by
    \begin{equation*}
        (\sum_{r=1}^R2^{R-r})-(m-1)=(\sum_{r=0}^{R-1}2^{r})-(m-1)=(2^R-1)-(m-1)=2^R-m.
    \end{equation*}
    \end{proof}
    
    \begin{defi}
        As a result of the previous process, we obtain a matrix $L$ with $m\times 3$ elements. Each row corresponds to one match/comparison, in the order in which they have been performed. Therefore, for the $i$-th pairwise comparison ($i=1,2,...,m-1$), the item $L_{i,1}$ contains the winner of the match $w(i)=1,...,m$, $L_{i,2}$ contains the loser $l(i)=1,...,m$, and $L_{i,3}$ contains the score, i.e., the number of units representing the difference of attractiveness $p_{L_{i,1}L_{i,2}}=p_{w(i),l(i)}\in \N\cup\{0\}$ of the winner over the looser. For simplicity, we adopt the convention $L_{m,1}=L_{m,2}=w(m)=l(m)$ and $p_{L_{m,1}L_{m,2}}=p_{w(m),l(m)}=0$. The matrix $L$ constructed in this way is called the match matrix associated with the TT problem.
    \end{defi}

   \begin{exam}
       To illustrate the process, we have asked a colleague to provide four of his favorite movies. She selected the following films: $a_1$- Back to the future, $a_2$- The Matrix, $a_3$- The Batman, and $a_4$- Rocky. To carry out the TT method, she suggests pairing $a_1$ and $a_2$, and $a_3$ and $a_4$ because she thinks that it is clearer. So, we start the questioning protocol by asking her: \textit{Which one do you prefer, $a_1$ or $a_2$?}. She replied that she preferred $a_1$. Then, we asked to represent the difference of attractiveness between $a_1$ and $a_2$ by using black cards. She decided to place $1$ card between them. From these questions, we deduced that the preference matrix $M$ associated with her evaluation would satisfy $M_{1,2}=(1+1)=2$. Afterwards, we followed a similar protocol to obtain the difference of atractiveness between $a_3$ and $a_4$. From her responses, we obtained $M_{3,4}=1$. Subsequently, since the winners of the first round are $a_1$ and $a_3$, she repeated the process for these two objects, deducing that $M_{1,3}=4$ and, consequently, that $a_1$ is the winner of the tournament. Therefore, we obtain the following match matrix:
       \begin{equation*}
           \begin{pmatrix}
               1&2&2\\
               3&4&1\\
               1&3&4
               \end{pmatrix}.
       \end{equation*}
   \end{exam}

\subsection{Stage II. Processing}
\noindent Now, let us describe how to derive the pairwise comparisons of all the objects using the information provided in the match matrix computed in the previous stage. Let us consider a TT problem to assess $m\in\N$ objects and let R=$C(\log_2(m))$ be the number of rounds. Consider the associated match matrix $L$.

\begin{itemize}
    \item[~] Step 1. Initialize a matrix $M$ with size $m\times m$.
    \item[~] Step 2. Complete the comparisons of the objects with themselves. Make $M_{i,i}:=0$ for $i=1,2,...,m$.
    \item[~] Step 3. Input the available information in the match matrix. For $i=1,2,...,m-1$, write 
    $M_{L_{i,1},L_{i,2}}:=L_{i,3}$ and $M_{L_{i,2},L_{i,1}}:=-L_{i,3}$.
    \item[~] Step 4. Obtain the evaluation of the winner against all the remaining objects. To do so, for $i=1,2,...,m-1$, write  $M_{L_{m-i,2},L_{m,1}}:=M_{L_{m-i,2},L_{m-i,1}}+M_{L_{m-i,1},L_{m,1}}$ and $M_{L_{m,1},L_{m-i,2}}:=-M_{L_{m-i,2},L_{m,1}}$.
    \item[~] Step 5. Complete the matrix using the winner. For $1\leq i<j\leq m$, if $M_{i,j}$ is empty, make $M_{i,j}:=M_{i,L_{m,1}}+M_{L_{m,1},j}$ and $M_{j,i}:-M_{i,j}$.
    [~]
\end{itemize}

\begin{defi}
    Given a TT problem to assess $m\in\N$ objects, the APR $M$ constructed following this process is called the preference matrix of the TT problem.
\end{defi}

It is clear that the above-described process is exhaustive, in the sense that, when it finishes, we obtain the values for all the possible pairwise comparisons. Additionally, the matrix $M$ obtained through this algorithm satisfies many interesting properties related to reciprocity and consistency. Let us start by showing that the evaluation matrix $M$ is indeed reciprocal.

\begin{prop}
     Given a TT problem to assess $m\in\N$ objects, the associated evaluation matrix $M$ is additively reciprocal.
\end{prop}
\begin{proof}
By construction, $M_{i,j}+M_{j,i}=0$ for all $i,j=1,...,m$.
\end{proof}

At this stage, let us show that the evaluation matrix constructed through this process is consistent. We start with the following result, which will be essential in the proof.

\begin{lem}\label{lemmaTTM}
    Let $M$ denote a preference matrix with dimension $m\times m$. Suppose that there exists $l=1,...,m$ such that $M_{i,j}=M_{i,l}+M_{l,j}$ for all $i,j=1,...,m$. Then $M$ is consistent.
\end{lem}
\begin{proof}
    Given $k=1,...,m$, the following holds:
    \begin{equation*}
            M_{i,k}+M_{k,j}=M_{i,l}+M_{l,k}+M_{k,l}+M_{l,j}=M_{i,l}+M_{l,j}=M_{i,j}.
    \end{equation*}
    Since $k=1,...,m$ is arbitrary, we have $M_{i,k}+M_{k,j}=M_{i,j}$, which is the consistency of M.
\end{proof}

\begin{theo}\label{theoTTM}
    The preference matrix constructed using the TT method is consistent.
\end{theo}
\begin{proof}
    Given a TT problem to assess $m\in\N$ objects, consider the match matrix $L$. Let us show that the preference matrix $M$ of the TT problem  satisfies:
    \begin{equation*}
        M_{i,j}=M_{i,L_{m,1}}+M_{L_{m,1},j} \;\forall i,j=1,...,m.
    \end{equation*}
    To do that, consider the following cases:
    \begin{enumerate}
        \item If $i=L_{m,1}$, then $M_{i,j}=M_{L_{m,1},j}=M_{L_{m,1},L_{m,1}}+M_{L_{m,1},j}$.
        \item If $j=L_{m,1}$, then $M_{i,j}=M_{i,L_{m,1}}=M_{i,L_{m,1}}+M_{L_{m,1},L_{m,1}}$.
        \item If $i,j\neq L_{m,1}$, then we have three possibilites:
        \begin{itemize}[label={--}]
            \item The item $M_{i,j}$ has been assessed in Step 2. Then $i=j$ and $M_{i,i}=0=M_{i,L_{m,1}}+M_{L_{m,1},i}$.
            \item The item $M_{i,j}$ has been assessed in Step 3. Let us suppose that such a comparison is the $k$-th one. Also, assume that the winner was $j$ and the loser $i$, i.e., $j=L_{m-k,1}$ and $i=L_{m-k,2}$. Then, when applying Step 4, we have
            \begin{equation*}
                M_{L_{m-k,2},L_{m,1}}=M_{L_{m-k,2},L_{m-k,1}}+M_{L_{m-k,1},L_{m,1}},   
            \end{equation*}
            which means 
            \begin{equation*}                
             M_{i,L_{m,1}}=M_{i,j}+M_{j,L_{m,1}},
            \end{equation*}
            and, therefore, $M_{i,j}=M_{i,L_{m,1}}-M_{j,L_{m,1}}=M_{i,L_{m,1}}+M_{L_{m,1},j}$. The case  $i=L_{m-k,1}$ and $j=L_{m-k,2}$ is similar.
            \item The item $M_{i,j}$ has been assessed in Step 5. In such case, by definition $M_{i,j}=M_{i,L_{m,1}}+M_{L_{m,1},j}$.
        \end{itemize}
    \end{enumerate}
    Since $M_{i,j}=M_{i,L_{m,1}}+M_{L_{m,1},j} \;\forall i,j=1,...,m$, then we can apply the previous lemma with $l=L_{m,1}$ and conclude that $M$ is consistent.
\end{proof}

We have seen that, when the decision-maker provides the pairwise comparisons by following the TT framework, we can apply our processing scheme to obtain a consistent preference matrix. Following the ideas developed in \cite{IVPR}, we can guarantee that, if the decision-makers provide cards representing the ratio, we can also apply the TT method to obtain a consistent multiplicative preference relation. 

\begin{exam}
    Let us continue with the example shown in the previous section. After applying Steps 1,2 and 3, we obtain the matrix
    \begin{equation*}
           \begin{pmatrix}
               0&2&4&-\\
               -2&0&-&-\\
               -4&-&0&1\\
               -&-&-1&0
           \end{pmatrix}.
       \end{equation*}
        Subsequently, we compute the remaining preferences of the best one, $a_1$, against the others, in this case only $a_4$. Therefore, $M_{1,4}=M_{1,3}+M_{3,4}=5$ and $M_{4,1}=-M_{1,4}=-5$. Thus, we obtain the following matrix:
         \begin{equation*}
           \begin{pmatrix}
               0&2&4&5\\
               -2&0&-&-\\
               -4&-&0&1\\
               -5&-&-1&0
           \end{pmatrix}.
       \end{equation*}
        Finally, we compute the remaining values using the information regarding the winner: $M_{2,3}=M_{2,1}+M_{1,3}=2$, $M_{3,2}=-M_{2,3}=-2$, $M_{2,4}=M_{2,1}+M_{1,4}=3$, and $M_{4,2}=-M_{2,4}=-3$, deducing that 
        \begin{equation*}
           \begin{pmatrix}
               0&2&4&5\\
               -2&0&2&3\\
               -4&-2&0&1\\
               -5&-3&-1&0
           \end{pmatrix}.
       \end{equation*}
\end{exam}

\subsection{Stage III. Evaluation}
\noindent In the previous steps, we have described how to retrieve the preferences from the decision-makers with a few questions and how to use the obtained information to build a consistent preference matrix. Now, we propose a method to derive a final score for each object from the preference matrix of the TT problem. Additionally, we justify that such a method can be applied even when the APR is not consistent. Let us start by making an observation regarding preference matrices.

\begin{theo}
    An preference matrix $M$ of size $m\times m$ is consistent if and only if there exist  $u_1,...,u_n\in\R$ such that $M_{i,j}=u_i-u_j\;\forall i,j=1,...,m$.
\end{theo}
\begin{proof}
    Suppose that $M$ is a consistent preference matrix. Then, let us define $$u=(M_{1,1},M_{2,1},...,M_{n,1}).$$ Since $M$ is consistent, we have
    \begin{equation*}
            M_{i,j}=M_{i,1}+M_{1,j}=u_i-u_j \;\forall i,j=1,...,m.      
    \end{equation*}
    Conversely, let us assume that $M_{i,j}=u_i-u_j\;\forall i,j=1,...,m$. In such case, for $i,j,k=1,...,m$,
    \begin{equation*}
        M_{i,k}+M_{k,j}=u_i-u_k+u_k-u_j=M_{i,j},
    \end{equation*}
    which is the consistency of $M$.
    \end{proof}

The values $u_1,...,u_n$ obtained in this way act as a value scale for the corresponding set of objects. In this sense, these values allow obtaining a global performance of each object for a consistent matrix. 
    
Finally, we show that, for a TT preference matrix, it is possible to obtain a normalized value scale. Note that the maximum possible difference between objects can be computed as $$\max_{k=1...n}M_{L_{m,1},k},$$ which is the number of units between the winner and the worst object of the tournament. Let us assume that the maximum is reached at the item $M_{L_{m,1},k^*}$, where $k^*$ is the index of the worst object. Then the values $M_{i,k^*}, i=1,...,n$ are exactly the number of units between the object $a_i$ and the worst object $a_{k^*}$. Thus, we may divide those by  $\max_{k=1...n}M_{L_{m,1},k}$ and obtain a normalized value scale whose maximum is $1$, minimum $0$, and each value is proportional to the number of cards placed between the corresponding object and the worst one. Let us summarize this in one Corollary.

    \begin{cor}
        Let $M$ denote a preference matrix constructed through the TTM for $n$ objects. Then, if $k^*$ denotes the index of the worst object, the values
        \begin{equation*}
            v_i=\frac{1}{M_{L_{m,1},k^*}}M_{i,k^*}, i=1,...,n
        \end{equation*}
        determine a normalized value scale.
    \end{cor}
    
Once the final values have been obtained, the objects may be ranked. Furthermore, the number of units between the objects can easily be deduced from the $k^*$-th column of the matrix $M$. For this reason, the results can be presented to the decision-maker according to the Deck of Cards method. In this sense, the objects will appear in an ordered way, and the corresponding number of cards can be placed by the analyst between the levels. In this way, the decision-maker could make further modifications if necessary, while the outputs could be easily computed in real-time.

    \begin{exam}
        Let us continue with our example. At this stage, we aim to obtain a final ranking for the objects, for which we will use the value scale obtained in the previous Corollary. The preference matrix was as follows:
        \begin{equation*}
           \begin{pmatrix}
               0&2&4&5  \\
               -2&0&2&3 \\
               -4&-2&0&1\\
               -5&-3&-1&0
           \end{pmatrix}.
       \end{equation*}
       Remember that the winner of the tournament was $a_1$. Therefore, we can identify the difference between the winner and the worst object at $M_{1,4}$, which implies that $a_4$ is the worst-positioned option. Consequently, a normalized value scale may be $v_1=5/5=1$, $v_2=3/5$, $v_3=1/5$ and $v_4=0$, which yields $a_4\succ a_2\succ a_3\succ a_4$.
    \end{exam}

\section{A web-based tool for TTM}\label{sec:app}

\noindent To illustrate the performance of the TTM, we have developed a web application (\href{http://suleiman.ujaen.es:8055/TTM/}{http://suleiman.\allowbreak ujaen.es:8055/TTM/}) that allows testing the performance and insights of TTM. This web-based tool also allows making modifications using the classical Deck of Cards method to show not only the compatibility between the two methodologies, but also the flexibility of TTM to be tuned by the decision-maker.

\begin{figure}[H]
    \centering
    \includegraphics[width=0.8\linewidth]{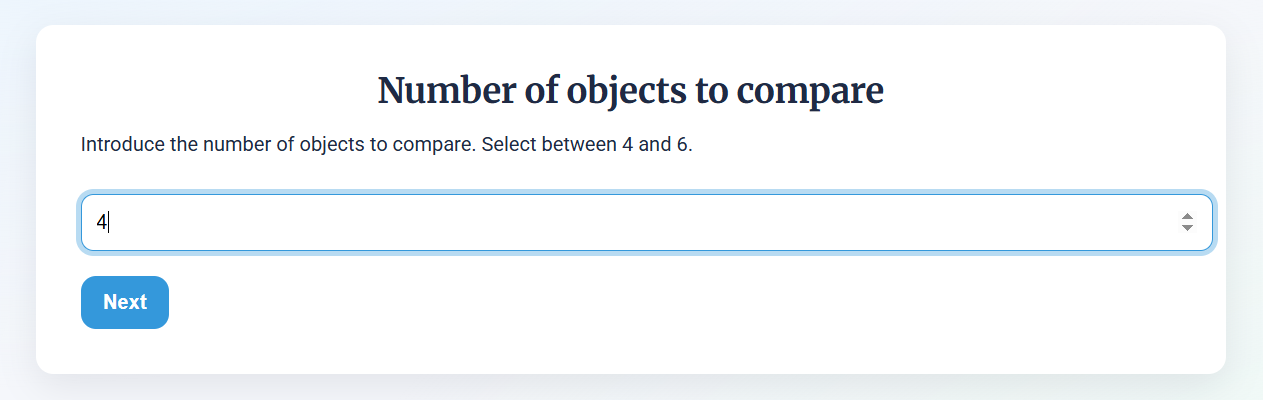}
    \caption{Step 1 of the TTM web-based tool.}
    \label{fig:ttmweb_step1}
\end{figure}

\noindent To facilitate the description of the tool, we reproduce the same situation that was previously described in the Examples of this manuscript. Therefore, let us assume that we ask the decision-maker to give us her opinion about her favorite movies and compare them with each other. The user can choose to compare 3 to 6 of their favorite movies. This first step is shown in Fig.  \ref{fig:ttmweb_step1}.

\begin{figure}[H]
    \centering
    \includegraphics[width=0.8\linewidth]{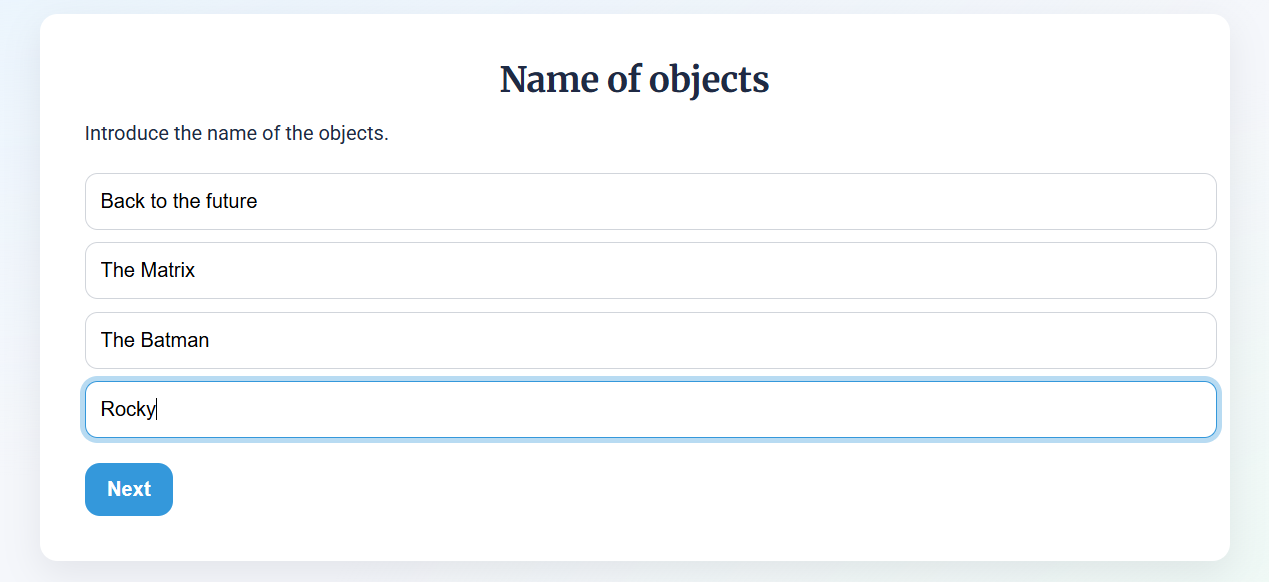}
    \caption{Step 2 of the TTM web-based tool.}
    \label{fig:ttmweb_step2}
\end{figure}

\noindent Once the number of objects has been set, the user has to provide the names of the objects that he/she wants to compare. Until the user enters the names of all the movies, she/he will not be able to continue with the next step. Furthermore, the user can go back to the previous step and change the number of alternatives to compare. In this case, our decision-maker wrote the names of her favorite movies, which are shown in Fig. \ref{fig:ttmweb_step2}.

\begin{figure}[H]
    \centering
    \includegraphics[width=\linewidth]{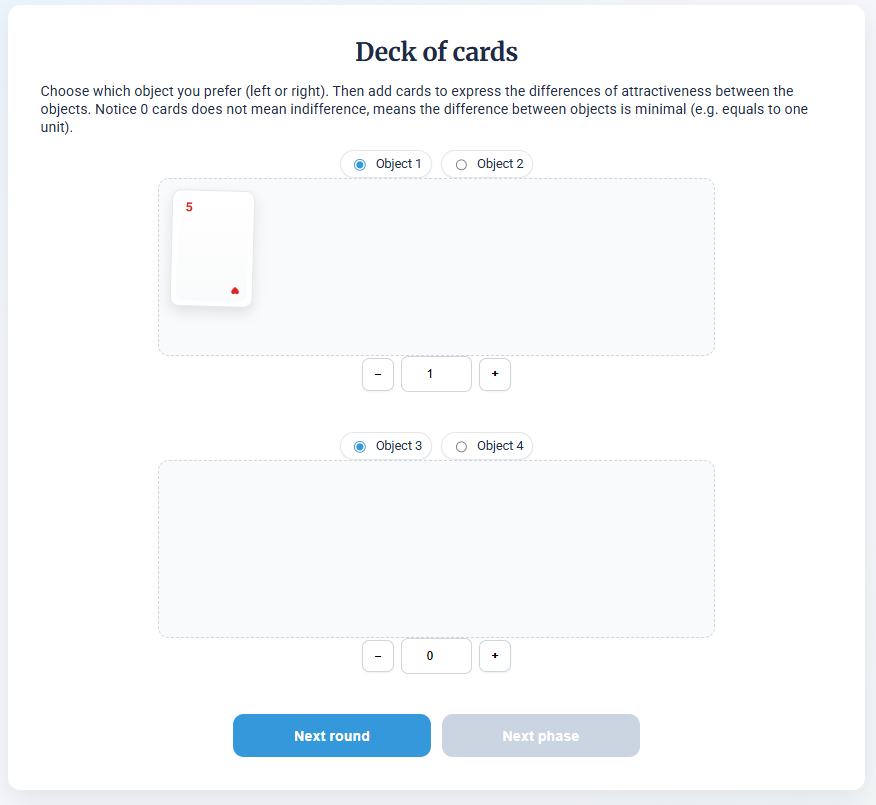}
    \caption{Step 3 (round 1) of the TTM web-based tool.}
    \label{fig:ttmweb_step3_r1}
\end{figure}

\begin{figure}[H]
    \centering
    \includegraphics[width=0.8\linewidth]{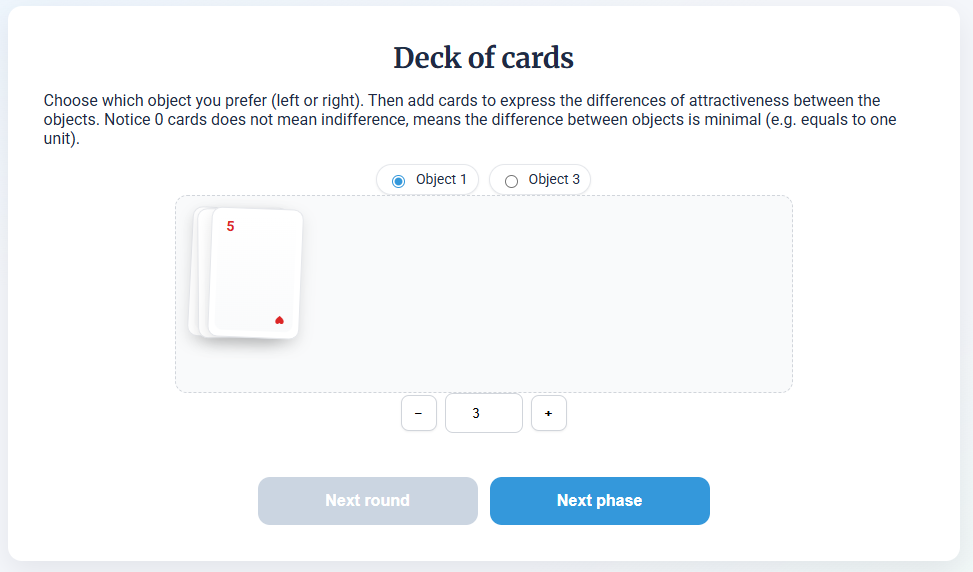}
    \caption{Step 3 (round 2) of the TTM web-based tool.}
    \label{fig:ttmweb_step3_r2}
\end{figure}

\noindent In the subsequent screen, the user will be asked to pairwise compare the objects according to the TTM method. To do so, several objects are displayed to the user, and he/she must click which one is preferred and how many cards he/she feels that better represent the difference in attractiveness between each pair of objects. Considering our example of 4 movies, the decision maker is initially asked to compare the first movie with the second and the third with the fourth (see Fig. \ref{fig:ttmweb_step3_r1}). Once these comparisons have been made, the user must click on the 'Next round' button. At that moment, and depending on the preference values entered by the user, the following confrontations will be shown. Following the example, the next, and last comparison would be between the alternatives 1 and 3 (see Fig. \ref{fig:ttmweb_step3_r2}). In case of a greater number of alternatives, the process will be the same until there are no more matches. 

\begin{figure}[H]
    \centering
    \includegraphics[width=0.8\linewidth]{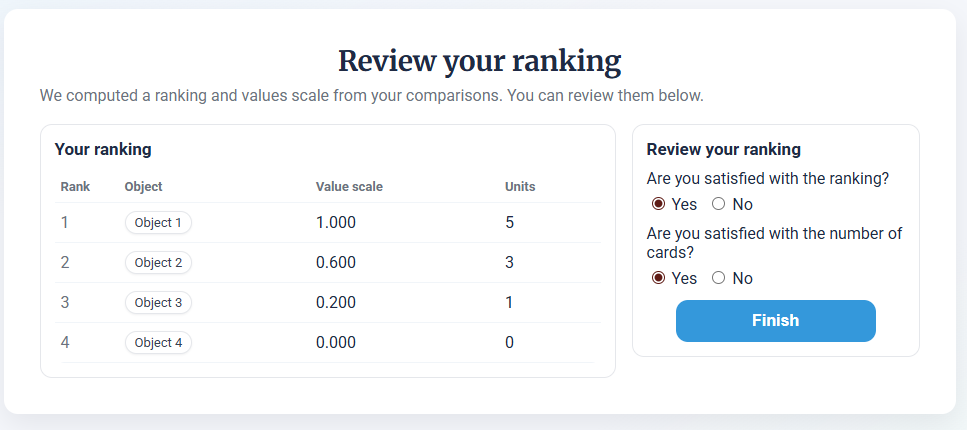}
    \caption{Step 4 of the TTM web-based tool.}
    \label{fig:ttmweb_step4}
\end{figure}

\begin{figure}[H]
    \centering
    \includegraphics[width=0.8\linewidth]{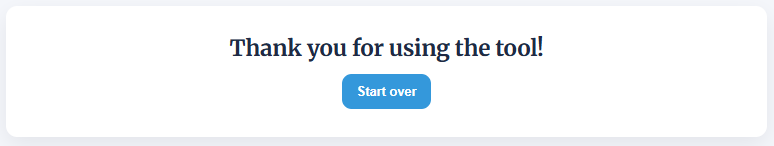}
    \caption{Final step.}
    \label{fig:ttmweb_finalstep}
\end{figure}

\noindent Once the tournament has finished, the results are shown to the user on a new screen. In the new window, the tool displays the ranking of the objects, the associated value scale, and the number of units associated with them (see Figure \ref{fig:ttmweb_step4}). In addition, the web-based tool allows the user to make modifications to his/her own preferences. He/she can modify both the ranking and the value scale provided by the TTM. If the decision maker is satisfied with both the ranking and the value scale, the process finishes, and the tool shows a farewell message (see Fig.~\ref{fig:ttmweb_finalstep}). If the user wishes, he/she can return to the initial step and start the process again.

\begin{figure}[H]
    \centering
    \includegraphics[width=0.8\linewidth]{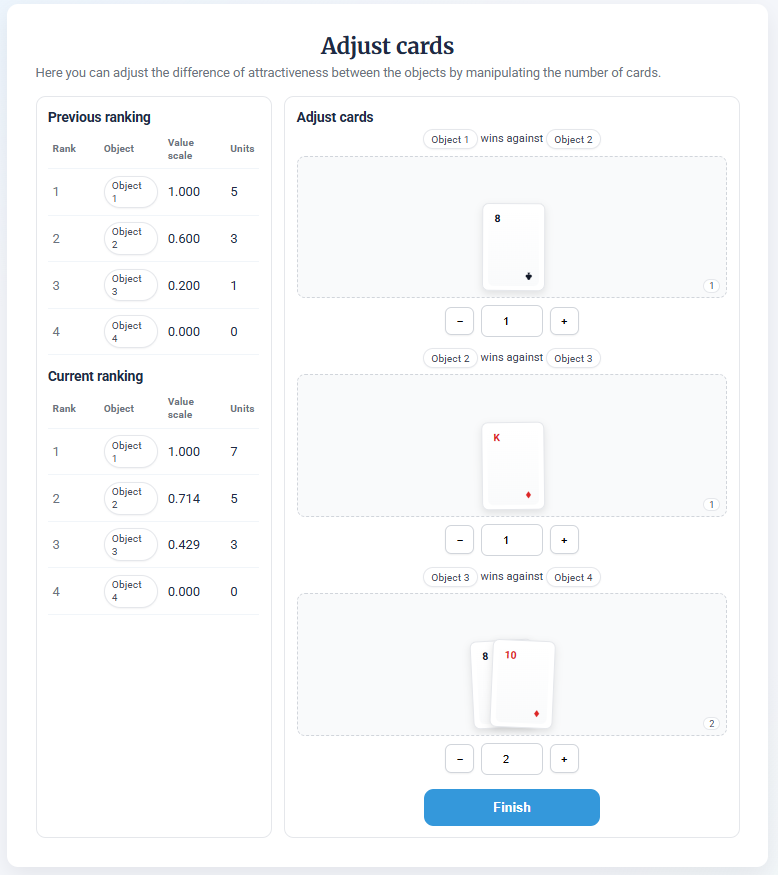}
    \caption{Step 5 of the TTM web-based tool in case the user agrees with the TTM ranking with the modifications done by the user.}
    \label{fig:ttmweb_step5_modi}
\end{figure}

In case the user is satisfied with the ranking, but wants to make modifications on the value scale, the tool will conduct he/she to a new screen in which he/she can modify the number of cards between the different objects, which will appear ordered according to the ranking provided by the TTM. The user can add or remove cards as desired according to the previously accepted ranking. As cards are added or removed, the results are updated and displayed in the lower left table (see Figure~\ref{fig:ttmweb_step5_modi}). When the user has finished making changes, they can click the finish button and will be directed to the screen shown in Fig.~\ref{fig:ttmweb_finalstep}. 



\begin{figure}[H]
    \centering
    \includegraphics[width=\linewidth]{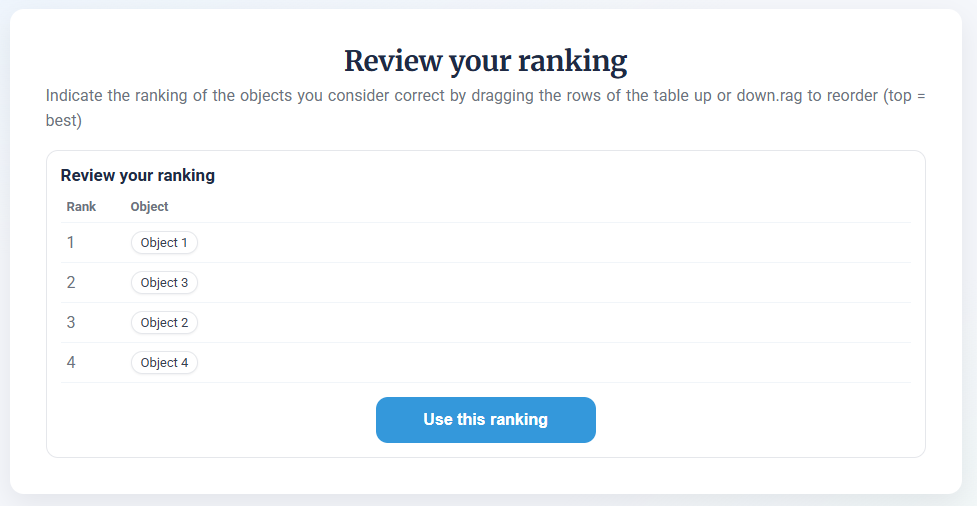}
    \caption{Step 4.1 of the TTM web-based tool (modified ranking).}
    \label{fig:ttmweb_step4_1_m}
\end{figure}

The other case to consider is that the user disagrees with the TTM ranking. In this case, the user is instructed to redefine the ranking in the next screen. In the screen shown in Fig.~\ref{fig:ttmweb_step4_1_m}, the user can drag the ranking items to the desired position and set their preferred ranking. In this case, we see how the user has swapped the positions of the top 2 and top 3. After he/she stablishes the ranking, the tool allows for introducing cards between the different objects to account for the difference in attractiveness between the objects and determining the value scale.


From here on, the process is the same as explained above. The user can add or remove cards as they wish and see the results in the lower left table (see Fig.~\ref{fig:ttmweb_step5_modi}). The only difference is that initially, the user will not be shown any results, since when defining a new ranking, there is no information about the difference in attractiveness between the elements, only their order.


\section{Conclusions}\label{sec:conclusion}
\noindent This paper has presented the TTM, a new elicitation mechanism that overcomes many of the limitations of other preference structures in the literature that are based on pairwise comparisons.

TTM extends the Deck of Cards Method while using it as a basis for the elicitation process \cite{corrente2021}. However, the main advantage of the new proposal is that the decision-maker does not need to know in advance the ranking of the objects to be compared. In a similar way to a sports tournament, the objects will be matched and, after the successive games, the final winner will be decided. This process, although more general, requires exactly the same number of comparisons (one less than the number of objects to be compared) as the classical Deck of Cards method. Further, as we show in our example, the two methodologies can be combined to enhance the elicitation process.

Additionally, our proposal presents several advantages in comparison with other methods based on pairwise comparisons. As we said before, the cognitive effort from the decision-maker is minimal, while he/she does not need to fill the full preference matrix. Furthermore, the values obtained through TTM are guaranteed to be consistent with the original information provided by the decision maker \cite{CAVALLO2025103175}. Further, the use of cards to measure the attractiveness between objects extremely facilitates the task to the decision-maker in terms of explainability, overcoming the limitations of Saaty's scale, in which it may not be clear the difference between assessing $7$ or $8$ to a certain pairwise comparison \cite{Saaty1990}.

From the computational point of view, our process simplifies all the models in the literature explicitly designed for manipulating preference matrices (AHP, FPRs). For example, in \cite{Rodriguez2021}, the authors proposed models for the value scale and also for the FPRs, concluding that the latter are much more time-consuming in computational terms. With our proposal, all these models are simplified since, after our elicitation process, the obtained value scale is completely consistent with the pairwise comparisons, or, in other words, they contain exactly the same information.

In future work, we will apply this proposal in contexts of missing information, as well as in real-world decision situations. Furthermore, we will work on further developing computational tools to simplify the accessibility of our methodology.

\section*{Acknowledgments}
\noindent José Rui Figueira is financed by Portuguese funds through the FCT – Foundation for Science and Technology under project UID/97/2025 (CEGIST). Diego García-Zamora is financed by the mobility grant CAS24/00249 from the Spanish Ministry of Science, Innovation, and Universities, which supported his research stay at the Instituto Superior Técnico, Universidad de Lisboa. Diego García-Zamora also acknowledges the support of CEGIST for all the assistance during his research stay.

\vfill\newpage

\begingroup
\setstretch{1.0}  
\bibliographystyle{model2-names}
\bibliography{biblio}
\endgroup
\end{document}